# Explainable Image Captioning using CNN- CNN architecture and Hierarchical Attention

Rishi Kesav Mohan, Vignesh S, Sanjay S.

*Department of Computer Science, College of Engineering, Anna University, Chennai, Tamil Nadu, India - 600025*

# Explainable Image Captioning using CNN- CNN architecture and Hierarchical Attention

**Abstract.** Image captioning is a technology that produces text-based descriptions for an image. Deep learning-based solutions built on top of feature recognition may very well serve the purpose. But as with any other machine learning solution, the user understanding in the process of caption generation is poor and the model does not provide any explanation for its predictions and hence the conventional methods are also referred to as Black-Box methods. Thus, an approach where the model's predictions are trusted by the user is needed to appreciate interoperability. Explainable AI is an approach where a conventional method is approached in a way that the model or the algorithm's predictions can be explainable and justifiable. Thus, this article tries to approach image captioning using Explainable AI such that the resulting captions generated by the model can be Explained and visualized. A newer architecture with a CNN decoder and hierarchical attention concept has been used to increase speed and accuracy of caption generation. Also, incorporating explainability to a model makes it more trustable when used in an application. The model is trained and evaluated using MSCOCO dataset and both quantitative and qualitative results are presented in this article.

**Keywords:** Explainable AI, Image captioning, CNN decoder, Hierarchical attention, MSCOCO dataset.

**Introduction**

The process of providing a textual description for an image by identifying the various actors and environment in an image is regarded as Image Captioning. Being a subfield of computer vision and Natural Language Processing, it is a computationally intensive application that requires a large amount of data to accommodate the various actors and actions in an Image.

Deep Learning models use Encoder - Decoder model where the encoder is CNN (Convolutional Neural Network) and the decoder is a RNN (Recurrent Neural Network) with an efficient proposal by K. Xu, J. Ba et al., [1].

The input image is given to CNN to extract the features. The last hidden state of the CNN is connected to the Decoder. The Decoder is a Recurrent Neural Network (RNN) which does language modelling up to the word level.

Various approaches have been proposed and papers have been published regarding the methods or approaches for image captioning. One such approach is using CNN+CNN network for caption generation where both encoder and the decoder use CNN instead of RNN. This article proposes a caption generator using the above approach which proposed faster training time and 3x better result compared to CNN-RNN network. This algorithm is derived from Qingzhong Wang and Antoni B. Chan [10] which proposes an approach to use CNN as the decoder instead of the traditionally used LSTM and Xinyu Xiao et al., [9] which proposed a Hierarchical CNN-RNN network as the architecture.

Though the caption generation provides good results and produces captions of understandable and appreciable quality, still we can only regard the model as a Black-Box model as the user or the practitioner of the generated model doesn't know how an output is generated for an Input by the model.

Thus, for all existing algorithms an explainable approach can be implemented so that the interoperability of the model and the user is enhanced and also the trust between a user of a model and the application is improved.

Ho-Jin Choi et al.,[14] used a Region-Word attention model to compute a relevance score between a region and a word identified by a model. It used a CNN-LSTM architecture as the caption generator. One primary drawback of LSTM is that it works using memory states. Input from previous cells has to be remembered to derive an output, thus eliminating the opportunity to introduce parallelism. On the other hand, a decoder using Causal CNN can be implemented having parallelism in mind making training faster while also maintaining the functionality of a LSTM. Thus, this article proposes Causal CNN as the decoder. The following contributions are made in this paper:

- It introduces a CNN-Causal CNN Encoder-Decoder model for caption generation which is faster and outperforms LSTM-based models.
- A hierarchical network is proposed for the attention module which forms a bridge between the Visual CNN that extracts the image features and the Language CNN which generates the captions.
- A Region-Word attention model to identify the various actors present in an image and presenting the relevance between an actor and an image.
- An Interpretability enhancement module that derives a loss based on the scores obtained from Region-Word attention module.

The rest of this paper is structured as following, the related works is presented in section 2; the proposed model is demonstrated in section 3; the evaluation and computational results are presented in section 4;

**Related Works**

Image Captioning is a task which is one of the primary goals of computer vision. Since serving as the combination of both computer vision and natural language processing it is also considered a difficult problem. But due to the presence of large classification datasets as used in Russakovsky et al., [2] visual representation of images is now possible.

But one of the important issues that existed with generating captions was the absence of any method to find out what the model actually "saw" before it is deriving any captions. K. Xu, J. Ba et al., [1] proposed an approach to generating captions where an "attention mechanism" was incorporated. It proposed two attention states, "soft" and "hard". Soft attention is an average weighted sum of all hidden states that caused an activation. Hard attention focused only on a single state with the highest score. The soft attention mechanism can be used to identify which region in an image caused the caption to identify it as an object. For all the operations or tasks, we used MSCOCO dataset as first used in Lin Tsung-Yi, Maire et al., [3]. The 2017 updated dataset consists of "80K" images to train from which would help us greatly as the captions generated are more proper as the number of training samples increases. Since captioning an image is similar to translating an image to a sentence, the most suited framework would be to use a "encoder-decoder" framework as proposed by Cho et al., [4]. The first neural networks-based approach to image captioning was done by Kiros et al.,[5] who proposed a multimodal log-bilinear model that was biased by features from the image. But they used a feed-forward neural language model which was later replaced with a recurrent model in Mao et al., [6].

For visual understanding of an image, via the encoder, most of these networks use the last convolutional layer of a network designed for some computer vision tasks. There are several downsides to that. First, these models are specialized to detect certain objects from the image. Thus, when we get deeper into the network, the network focuses on these objects, becoming almost blind to the rest of the image. These blind spots of the encoder sometimes are where the next word in the caption lies. Moreover, many words in the caption are not included in the target classes of these tasks, such as "snow". Having this observation in mind, Kelvin Xu, Jimmy Ba, Ryan Kiros, et al [1] proposes a method in order to reduce the blind spots of the last convolutional layer of the encoder, they propose a novel method to reuse other convolutional layers of the encoder. Doing so provides us diverse features of the image while not neglecting almost any part of the image and hence, we "attend to everything" in the image.

    The crucial decision before implementing captioning is choosing the convolutional feature extractor. The captions are generated using a CNN-CNN encoder-decoder model. Simonyan, K, Zisserman, A [7] proposes a VGG (Visual Geometry Group) which produces a boost in performance over using AlexNet. It proposes a VGG-16 model for the encoder for the full image, and the decoder generates the words using a feature vector. Q. You, H. Jin, Z. Wang, C. Fang, and J. Luo. [8] proposed an image caption generator that also provided a clear description of semantically important objects that are needed exactly when they are needed. In particular, the semantic attention model has the ability to attend to a semantically important concept or region of interest in an image. The capacity of a single-layer network to integrate encoder and decoder outputs and inputs respectively is limited for a very complex task like image captioning. Also the problem of "vertical Depth" arises thus Xinyu Xiao et al., [9] proposes a deep hierarchical encoder decoder network where

the functions of encoder and the decoder respectively are separated. RNN is used generally as a decoder (LSTM). But using only CNN+RNN was not enough as the various salient features were ignored when generating a word. Thus, to imitate the attention process of human being's attention mechanisms were introduced [1]. But using RNN for decoder also has its disadvantages. The RNNs have to be calculated step by step, which cannot be applied using parallel computing and also there is a long path between start and end. Thus, alternatively an approach was proposed by Qingzhong Wang and Antoni B. Chan [10] where a CNN+CNN framework was proposed also providing faster results compared to LSTM-based models.

While generating a working efficient model is important, evaluating a trained model is of utmost importance to estimate the efficiency and quality of predictions predicted by a model. To evaluate an Image captioning model, there are several metrics like BLEU, CIDER, METEOR, ROGUE that are used for a long time. Peter Anderson, Basura Fernando, Mark Johnson, Stephen Gould., [11] proposed the metric SPICE which is used extensively today. Extensive evaluations across a range of models and datasets indicate that SPICE captures human judgments over model-generated captions better than other automatic metrics (e.g., system-level correlation of 0.88 with human judgments on the MS COCO dataset, versus 0.43 for CIDEr and 0.53 for METEOR). Furthermore, SPICE can answer questions such as `which caption-generator best understands colors?' and `can caption-generators count?'. The leaderboards are calculated based on SPICE score as the major criteria nowadays.

To explain the results produced by the caption generator the project needs to identify the various objects present in the image and must find its relevance with the caption generated. Girshick, R, Donahue, J, Darrell, T, et al. [12] used an Object-detection algorithm for the efficient detection of an Object in an image which can later be used to verify the caption.

The purpose of the attention model is to assist the generation part in considering region information. Han, S, Choi, H. [13] proposes a concept of attention mechanism and the attention model generates a weight matrix for input regions and words. The weight matrix generated was used to find a relevance score which expressed how much a word and region are relevant.

The relevance score thus generated can be used to visualize the region and the word which had the most relevance score. Ho Jin Choi et al., [14] in their model used this score to visualize the various objects identified in an image and visualize the corresponding word.

**Proposed Model**

*Model Architecture*

We use a two-component architecture to achieve the proposal. The first part generates captions with the image as the input and the second part to provide an explanation to it. The Interpretability loss obtained from the second component also is used to improve the quality of the first component. To identify the various Regions of Interest present in an image, a pretrained Object Detection algorithm implemented using Mask-RCNN algorithm is used. The complete model architecture with the two components high lightened is shown in Fig. 1 with the two rounded circles representing the output.

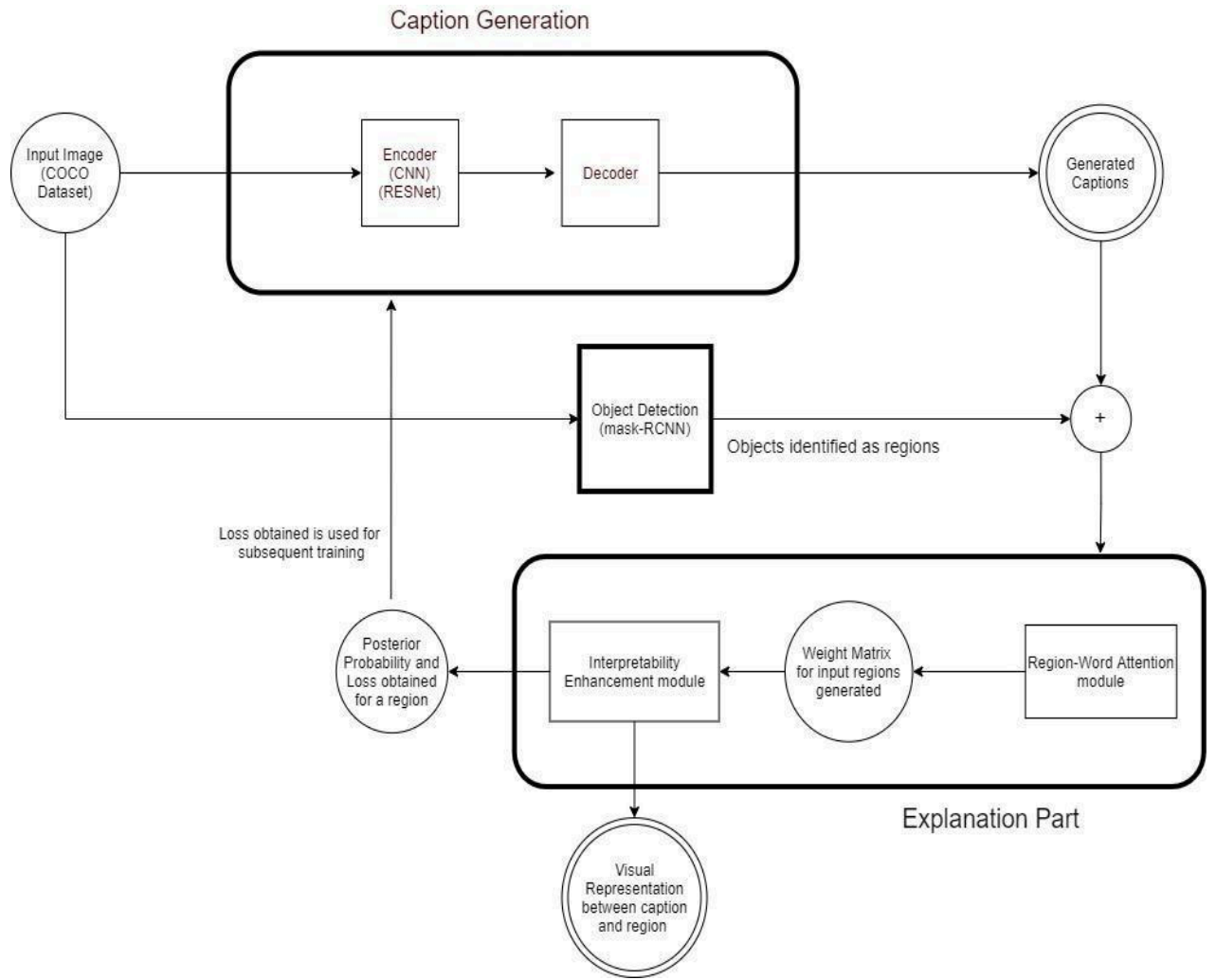

**Fig. 1.** Complete Model Architecture

*Caption Generation*

The caption generation part is used to generate the best possible caption for a given image. Instead of a conventional CNN-RNN encoder decoder pair we use a CNN-CNN encoder decoder pair to generate the caption. We also utilize hierarchical attention, so as to give importance to important features and have a higher impact on generated words in the caption. The attention network gives higher weightage to important features while neglecting insignificant ones.

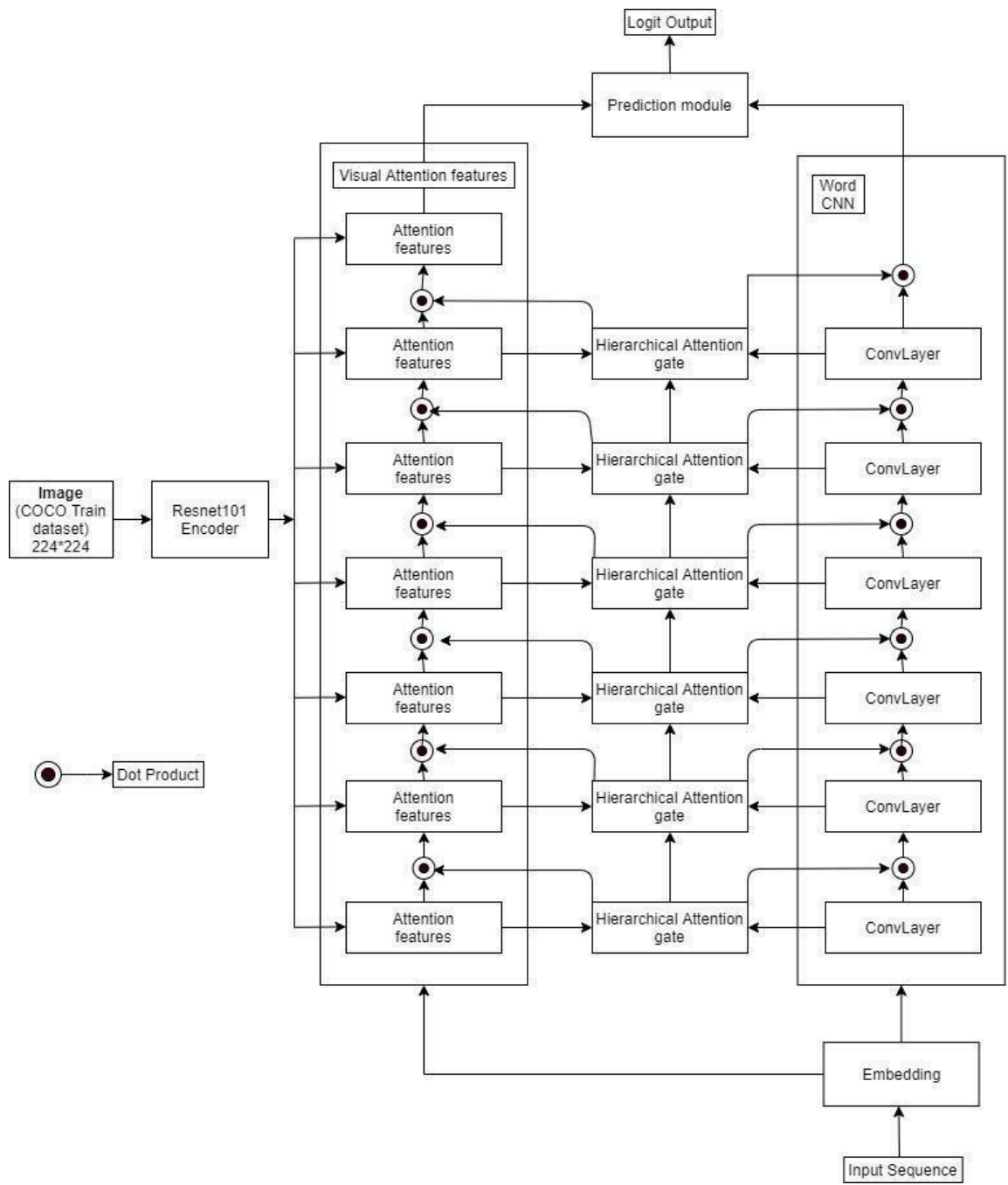

**Fig. 2.** Caption Generation Model

It consists of a stack of encoder - decoder architecture. The encoder used is a ResNet101 pre-weighted model. The encoder's visual feature map is shifted to the important features by using attention layers. The word-to-word concept mapping is extracted bottom up from the embedding layer through the decoder CNN layers. The CNN layers are Causal 1D Convolution layers. The advantage of using Causal Convolution layers over RNNs is that they perform faster than RNNs and allow feature maps to be cached between layers. In essence, we can eliminate the time penalty that comes with the usage of RNNs by substituting them with Causal CNNs. Between the encoder and decoder of each layer lie another hierarchical attention layer that passes the attention map onto the next layer. Each Hierarchical attention gate is a combination of 6 GRU cells. Feature map passed on to each higher layer is a dot map of the previous layer feature map and previous layer hierarchical attention map. An embedding layer at the lower level embeds the input sequence at the bottom level. A prediction layer at the top produce's logits for the predicted sentence. And finally, the logit output is converted into a sentence by referring to the dictionary which in our implementation had a vocabulary size of 9489 unique words.

*Explanation*

This part tries to provide an explanation for the generated caption and improve the quality of generated captions. At the same time a third outer component - the object detection model finds and isolates regions of interest in the image. Two components work to achieve explanation. The region - word attention model produces a probability score for each region word pair from object detection and the interpretability enhancement module improves the quality of generated captions. The Region-Word pair with the highest score is visualized using color coded boxes so that the user of our model can identify the various actors present in an image.

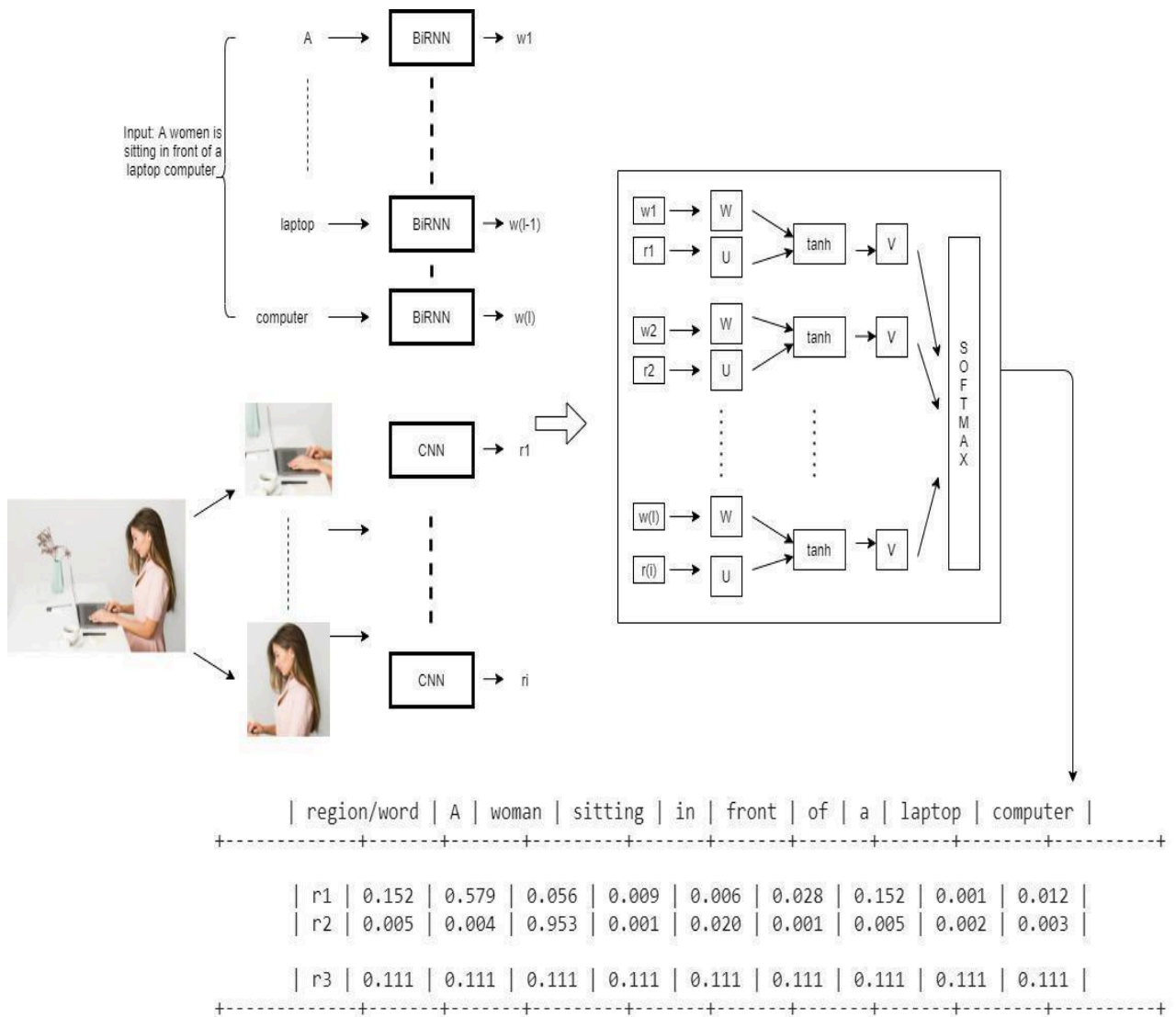

**Fig. 3.** Region – Word Attention model

*Region word attention model*

The purpose of the attention model is to assist the generation part in considering region information. To achieve this, we use a concept of attention mechanism and our attention model generates a weight matrix for input regions and words. Fig 4 depicts an example of the weight matrix obtained denoting the relevance score between the regions and the words. The attention model is parameterized as a feed-forward neural network, similar to other attention models.

We use a RNN-CNN pair to vectorize input sentences (from generation model) and regions of interest (from object detection). The RNN used is a bidirectional RNN. The CNN used is a pre-weighted Resnet-101. An activation layer employing tanh function is used to combine both the vectors and is softmaxed to obtain the resulting probability table.

```
+------------+-------+-------+---------+-------+-------+-------+-------+--------+----------+
| region/word|   A   | woman | sitting |  in   | front |  of   |   a   | laptop | computer |
+------------+-------+-------+---------+-------+-------+-------+-------+--------+----------+
|     r1     | 0.152 | 0.579 |  0.056  | 0.009 | 0.006 | 0.028 | 0.152 | 0.001  |  0.012   |
|     r2     | 0.005 | 0.004 |  0.953  | 0.001 | 0.020 | 0.001 | 0.005 | 0.002  |  0.003   |
|     r3     | 0.111 | 0.111 |  0.111  | 0.111 | 0.111 | 0.111 | 0.111 | 0.111  |  0.111   |
+------------+-------+-------+---------+-------+-------+-------+-------+--------+----------+
```

**Fig. 4.** Table representing the relevance scores obtained between the regions identified and the input caption

*Interpretability Enhancement module (IE)*

IE module helps us improve the quality of generated captions by taking region word attention into consideration. It determines whether the generator considers region word attention features while generating captions. The module selects those pairs with significantly higher relevance probability. Using these pair probabilities the loss for a given image and generated caption is calculated using Bayesian inference.

$$\text{Loss}_e = \sum_{i=1}^{n} \sum_{j=1}^{k} (1 - P(r_i|w_j))$$

$$P(r_i|w_j) = P(w_j|r_i) \times P(r_i) \tag{1}$$

*Visualizer*

The visualizer generates color coded sentences and bounding boxes to visualize the explanation generated. It generates the visualization only for those word-region pairs which have values above a dynamic threshold. This visualization helps the user to identify the actors present in a image more efficiently and the bounding boxes can also be used by other applications that use our model.

**Implementation**

Implementing the whole functional model involves two main steps. First we need to train a region - word attention model that will generate region word scores for caption generated and then fine tune the enhancement module and then use these to train and improve the caption generation model.

*Training the explanation part*

The dataset required for the explanation part is extracted from COCO dataset. The region word attention model is first trained to produce region word relevance score. Fig 3 presents the architecture diagram of the region-word attention model. The attention model is parameterized as a feed-forward neural network, similar to other attention models. It employs a Resnet-101 and a Bi-RNN in combination to vectorize sentence words (from generation) and regions (from object detection). The generated vectors are weighted with trainable parameters u and w and tanh value is calculated. It is scaled with another trainable parameter v and the resultant vector is softmaxed. This is done for every region-word pair to obtain the region-word probability. Then using this the interpretability module is fine tuned to select those pairs having the highest probability scores. From this we calculate the posterior probability using Bayesian inference and then the IE Loss.

*Training the generation part*

The generation part is trained initially to produce good captions without utilizing interpretability loss. This partially trained model is then retrained using self-loss along with IE Loss to improve the quality of generated captions. ResNet101 is used as the encoder. Features from the encoder are fed into n attention layers each having 2048 visual channels and 300 concept channels. These layers provide multi headed attention feedback at each level on visual and conceptual levels. An Embedding layer with a dictionary size of 9489 produces a vector of size 300. The decoder array is a stack of 6 1D Causal CNN with kernel size 3, which processes the vector produced by the embedding. Between the attention stack and the decoder stack lies the Hierarchical Attention stack which propagates the combination of both conceptual and visual attention upwards. It contains GRU cells of size 2348 and a hidden size 512. Combined with a pair of 1D CNN they produce the visual channel and conceptual channel output needed to be carried to the next layer. The topmost layer is the Prediction layer which is a combination of a pair of Dropout layer and 1D Convolution layer with a ELU activation unit in between. This layer produces the logits output required for sentence formation. The loss function used here is Cross Entropy Loss since the output produced is logits. The model contains about 11,84,66,961 trainable parameters out of which 4,46,54,504 are of ResNet parameters. Feedback loss along with self-loss from generation part is used to improve interpretability of generated captions. Fig 5 depicts the loss values obtained when the model was trained.

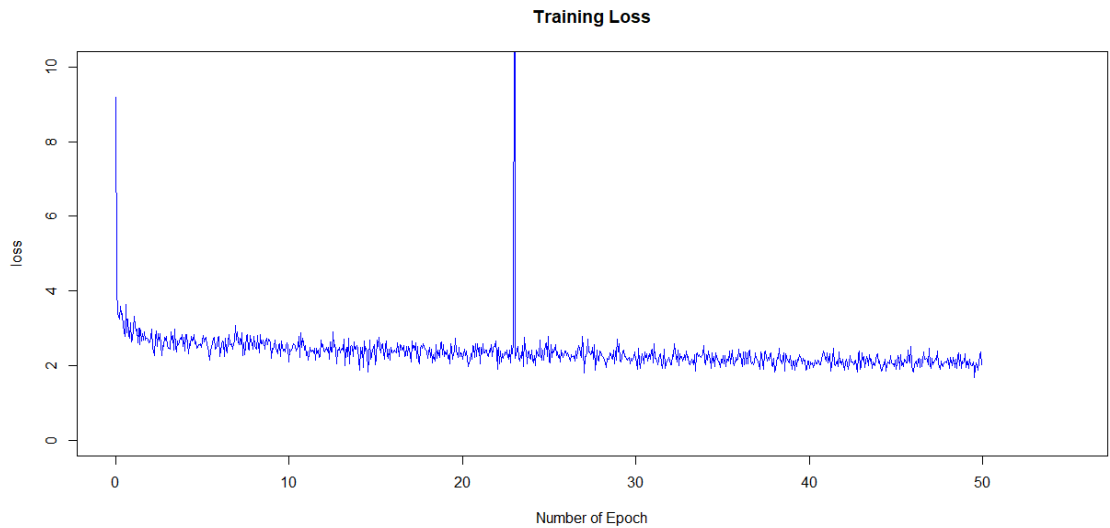

**Fig. 5.** Graph depicting the loss obtained with epochs for our model

**Results**

Image Captioning has shown progressive evolution over the years by scientists and researchers all over the world. This has also shown that there are many more improvements to be followed in the upcoming years. Our efforts lie primarily in using the best performing model and following it with consecutive explainability models using Explainable AI.

Initially, we evaluated our model with the standard set of evaluation metrics such as BLEU (Bilingual Evaluation Understudy), METEOR (Metric for Evaluation of Translation with Explicit Ordering), ROUGE (Recall-Oriented Understudy for Gisting Evaluation), CIDEr (Consensus-based Image De- scription Evaluation), SPICE (Semantic Propositional Image Caption Evaluation). Particularly, we have taken 4 orders of precision (N-gram) in computing BLEU score. We benchmarked our model with other models and our model showed relatively better performance compared to the various models present in the leaderboard. The scores for our model before the Explanation part is given below,

| Model | BLEU @1 | BLEU @2 | BLEU @3 | BLEU @4 | Meteor | Rouge | Cider | Spice |
|---|---|---|---|---|---|---|---|---|
| Hieratt CNN-CNN | 72.9 | 56.0 | 42.1 | 31.7 | 25.7 | 53.6 | 99.3 | 18.6 |

**Fig. 6.** Table denoting the values of evaluation obtained before IE

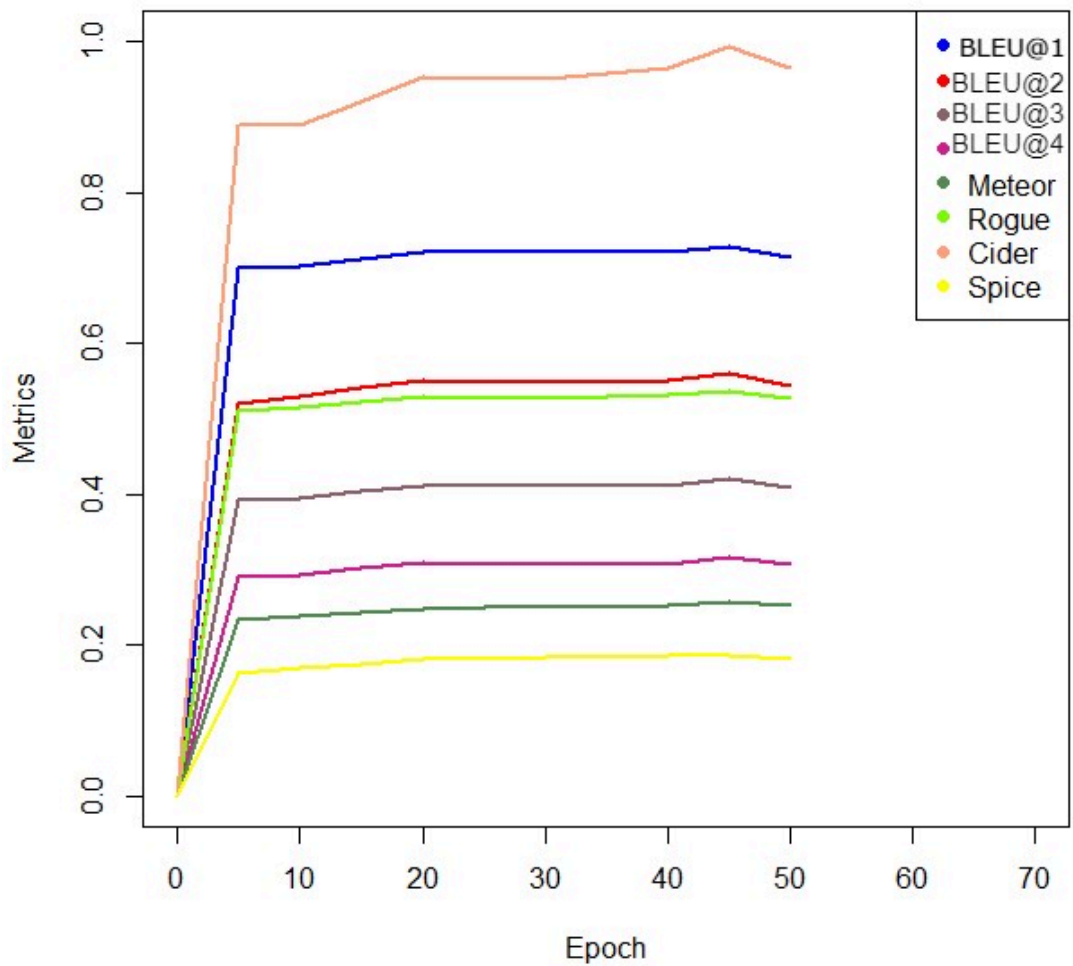

**Fig. 7.** Graph depicting the scores obtained for the various epochs before IE

Although our primary focus lies in interpreting and explaining the model to end users, we wanted them to be as accurate as possible. So, the calculated loss is further fed back to the model and helps improve the model using pre-calculated loss and the original input combined. Our Scores after Interpretability Enhancement model is given below

| Model | BLEU@1 | BLEU@2 | BLEU@3 | BLEU@4 | Meteor | Rouge | Cider | Spice |
|---|---|---|---|---|---|---|---|---|
| Hieratt CNN-CNN with | 72.7 | 55.7 | 41.8 | 31.6 | 25.8 | 53.7 | 99.8 | 18.9 |

| IE | | | | | | | | |
|---|---|---|---|---|---|---|---|---|

**Fig. 8.** Evaluation metrics obtained after IE

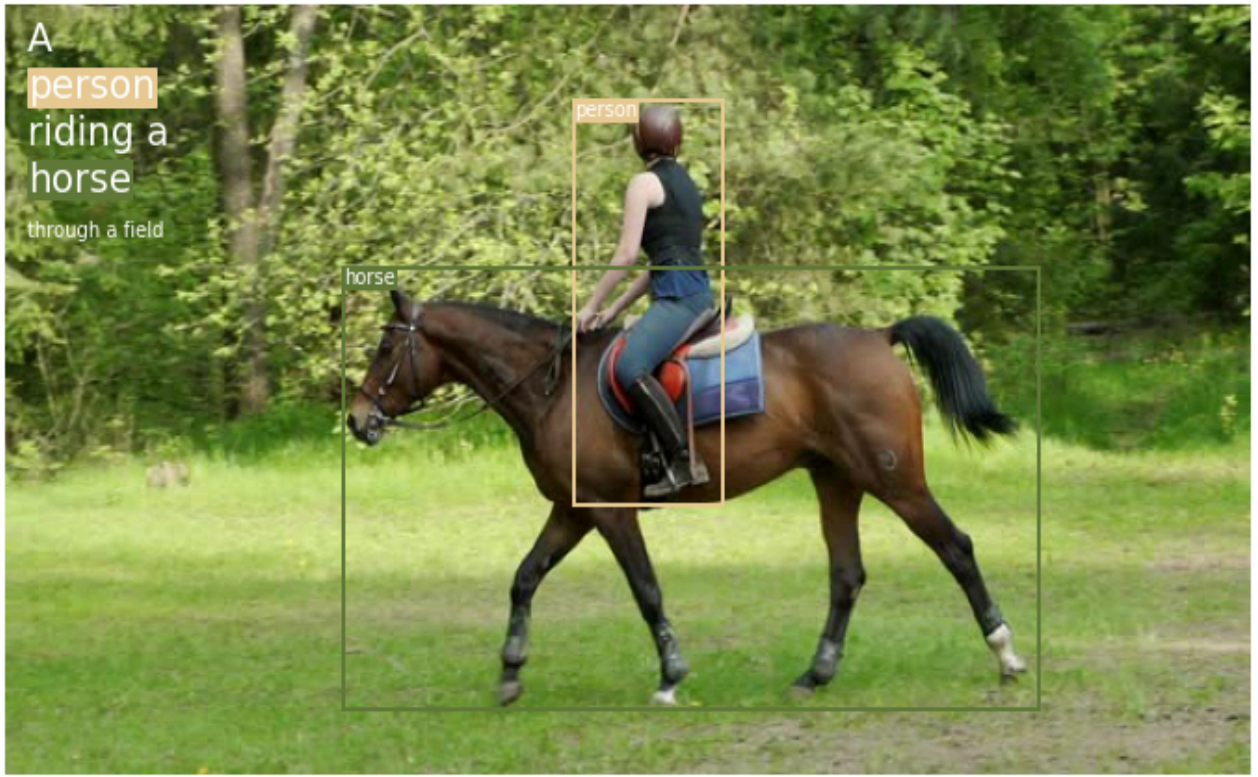

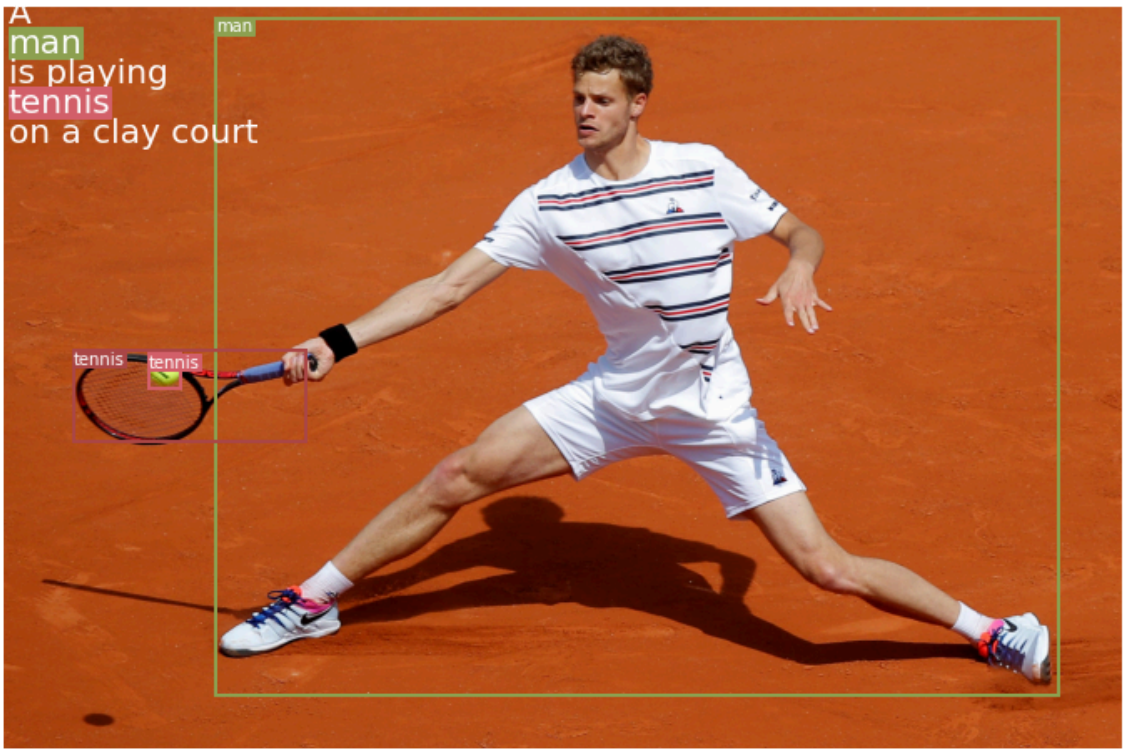

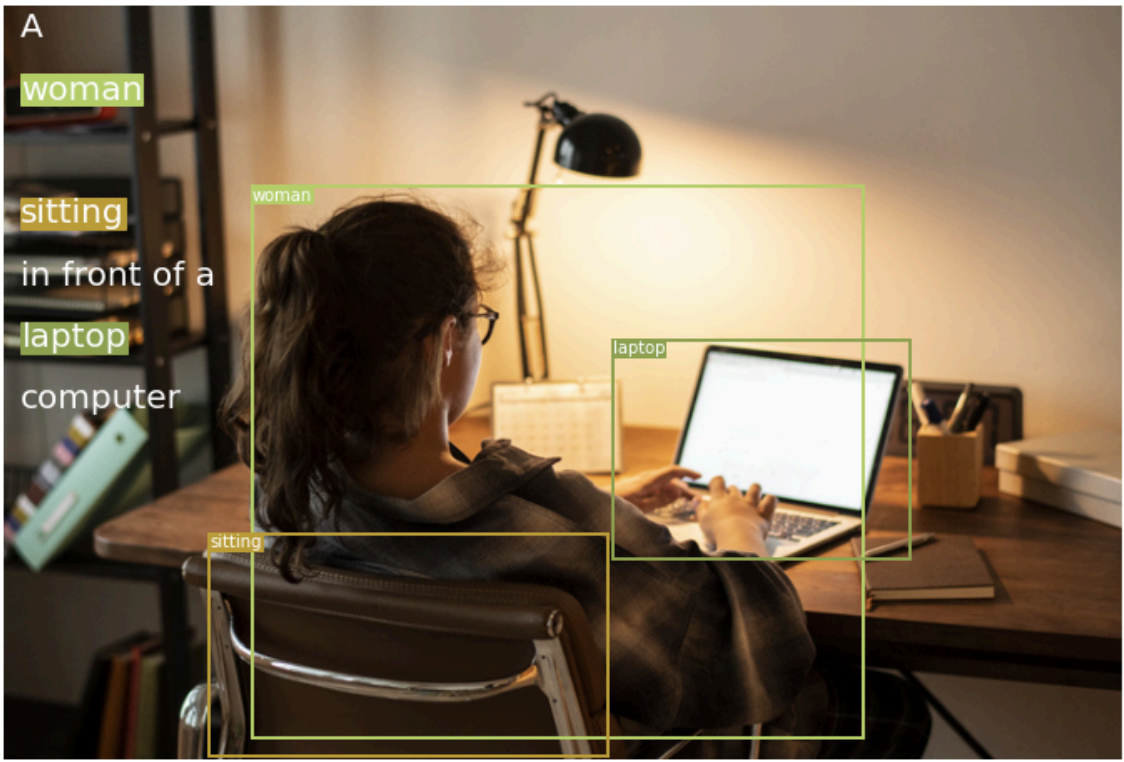

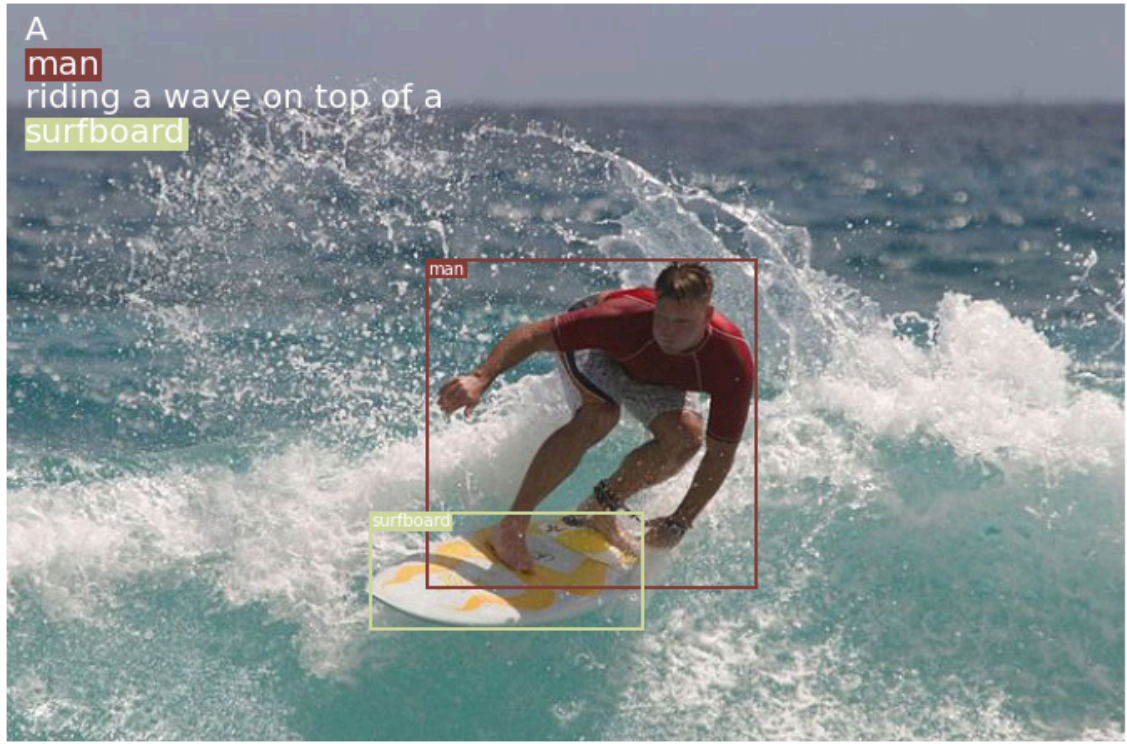

**Fig 9.** Captions generated by our model with various actors identified using color Boxes

**Conclusion**

In this paper, the model has been proposed and proper test cases and metrics have been identified and achieved. The various performance metric scores have been identified for our model using the evaluator provided for the COCO dataset and comparable scores have been obtained. We have finally been able to provide a way to make image captioning explainable by identifying the various actors and objects present in an image and also by identifying which word in the caption contributed to a region identified. Also, the loss obtained after interpretability enhancement would also be helpful in improving the model further. There is still space for larger future developments which include removing the false negatives that were identified, further improving explainable part to make the captions more explainable both in syntactic and semantic way.